# Vision transformer-based multi-camera multi-object tracking framework for dairy cow monitoring

Kumail Abbas[1,2], Zeeshan Afzal[3], Aqeel Raza[2], Taha Mansouri[3], Andrew W. Dowsey[4*], Chaidate Inchaisri[2*], Ali Alameer[3*]

[1] International Graduate Program of Veterinary Science and Technology, Faculty of Veterinary Science, Chulalongkorn University, Bangkok, 10440, Thailand.

[2] Research Unit of Data Innovation for Livestock Development, Department of Veterinary Medicine, Faculty of Veterinary Science, Chulalongkorn University, Bangkok, 10330, Thailand.

[3] School of Science, Engineering and Environment, University of Salford, United Kingdom.

[4] Bristol Veterinary School, University of Bristol, United Kingdom.

* Correspondence authors: andrew.dowsey@bristol.ac.uk, Chaidate.I@chula.ac.th, a.alameer1@salford.ac.uk

Kumail Abbas

drkumail.abbas@yahoo.com

https://orcid.org/0000-0003-4055-0915

Zeeshan Afzal

izeeshanafzal@outlook.com

https://orcid.org/my-orcid?orcid=0009-0001-0418-4121

Aqeel Raza

aqeel.r@chula.ac.th

https://orcid.org/0000-0002-0306-6967

Taha Mansouri

t.mansouri@salford.ac.uk

https://orcid.org/0000-0003-1539-5546

Andrew W. Dowsey

andrew.dowsey@bristol.ac.uk

https://orcid.org/0000-0002-7404-9128

Chaidate Inchaisri

chaidate.i@chula.ac.th

https://orcid.org/0000-0001-5940-2336

Ali Alameer

a.alameer1@salford.ac.uk

https://orcid.org/0000-0002-7969-3609

**HIGHLIGHTS**

- Real-time dairy cow monitoring using multi-camera homography-based panoramic view
- Vision Transformer model SAMURAI enhances segmentation with motion-aware memory
- YOLO11-m detector achieved mAP@0.5 of 0.97 and F1 score of 0.95 on test images
- Proposed system achieved 99.3% tracking accuracy and 99.7% IDF1 in complex barns
- Significantly outperforms Deep SORT in cow tracking, with minimal ID switches

**ABSTRACT**

Activity and behaviour correlate with dairy cow health and welfare, making continual and accurate monitoring crucial for disease identification and farm productivity. Manual observation and frequent assessments are laborious and inconsistent for activity monitoring. In this study, we developed a unique multi-camera, real-time tracking system for indoor-housed Holstein Friesian dairy cows. This technology uses cutting-edge computer vision techniques, including instance segmentation and tracking algorithms to monitor cow activity seamlessly and accurately. An integrated top-down barn panorama was created by geometrically aligning six camera feeds using homographic transformations. The detection phase used a refined YOLO11-m model trained on an overhead cow dataset, obtaining high accuracy (mAP\@0.50 = 0.97, F1 = 0.95). SAMURAI, an upgraded Segment Anything Model 2.1, generated pixel-precise cow masks for instance segmentation utilizing zero-shot learning and motion-aware memory. Even with occlusion and fluctuating posture, a motion-aware Linear Kalman filter and IoU-based data association reliably identified cows over time for object tracking. The proposed system significantly outperformed Deep SORT Realtime. Multi-Object Tracking Accuracy (MOTA) was 98.7% and 99.3% in two benchmark video sequences, with IDF1 scores above 99% and near-zero identity switches. This unified multi-camera system can track dairy cows in complex interior surroundings in real time, according to our data. The system reduces redundant detections across overlapping cameras, maintains continuity as cows move between viewpoints, with the aim of improving early sickness prediction through activity quantification and behavioural classification.

**Graphical Abstract**

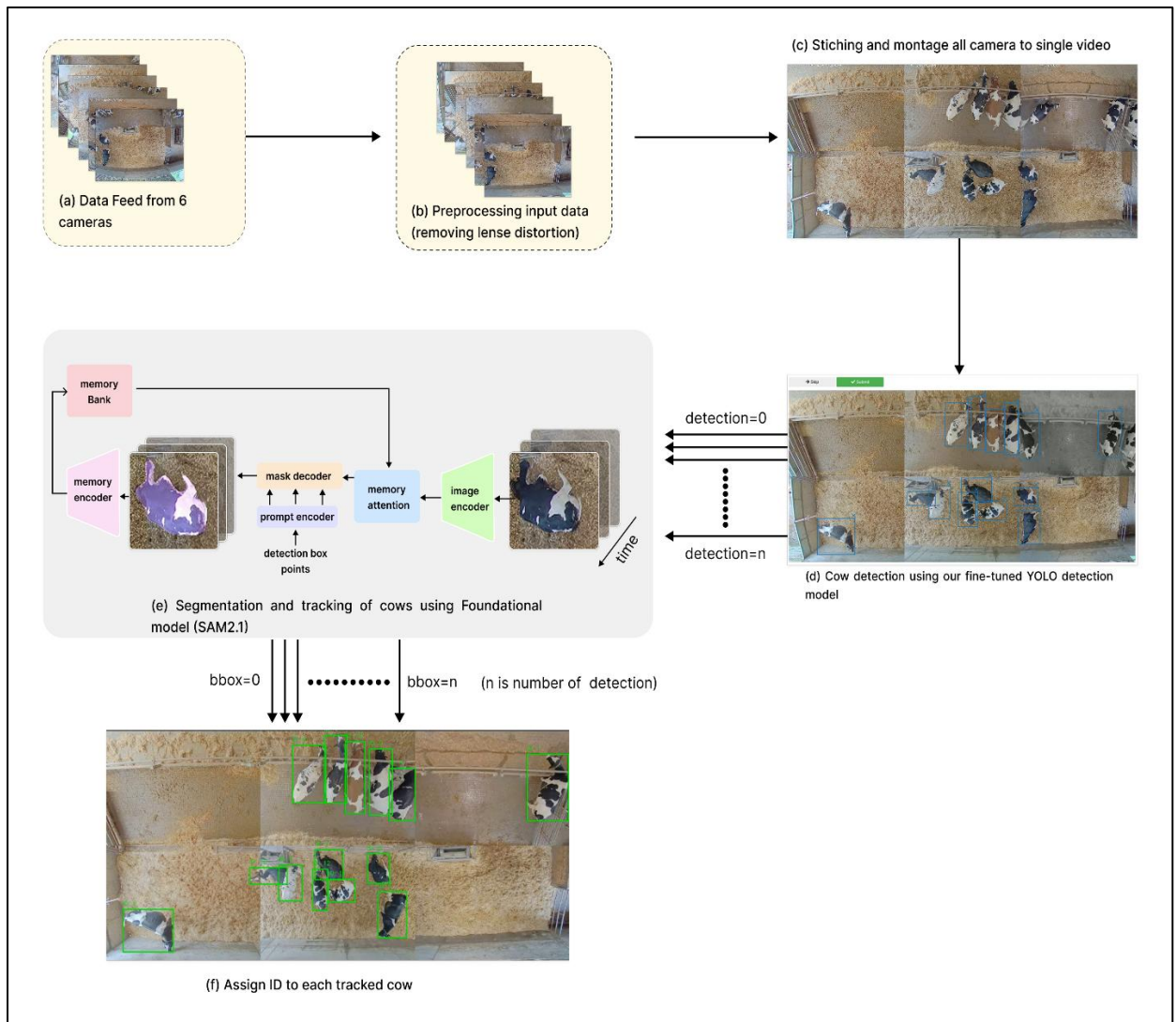

**Keywords:** Cow behaviour, Machine vision, Multi-camera tracking, Precision livestock farming

1. Introduction

Dairy farming is a foundational element of global agriculture, vital for providing a sizeable share of the world's milk supply and making a significant economic contribution. The global dairy industry is a multi-billion-dollar enterprise, with an estimated milking herd of over 942 million cattle producing approximately 965.7 million tonnes of milk in 2023 [1]. The longevity and productivity of dairy operations are directly correlated with the health and welfare of dairy cows [2]. Understanding the behavioural patterns of dairy cows is therefore paramount, as an animal behaviour is described as the patterns of observable, deliberate, or involuntary movements [3]. Cows tracking is essential in this context for monitoring and documenting the movement, location, and identification of individual animals within a herd [4]. The integration of consistent livestock tracking systems enables farmers to closely monitor the health of individual animals, identify irregular patterns, and manage diseases with greater effectiveness. This proactive strategy improves animal wellbeing during their life cycle and is useful in alleviating the economic losses linked to disease outbreaks [5].

Traditionally, the monitoring of dairy cow behaviour and health has relied on labour-intensive, and episodic methods, such as visual observations, physical examinations, and periodic veterinary assessments [6, 7]. The potential to reduce manual labour and improve management efficiency lies in the ability to recognize individual cow behaviour. The development and utilization of real-time tracking technologies have emerged as a promising tool for comprehensive dairy farm management, enabling optimal monitoring of individual animals and thereby improving the animal health and production outcomes through the use of sensing technology [8, 9]. Contact sensors like pedometers, Radio Frequency Identification (RFID) tags, and Inertial Measurement Units (IMUs) with a combination of accelerometers, gyroscopes and magnetometers, have been developed as wearable devices such as ear tags [10-12] and neck collars [13], to capture various behavioural movements and identify and monitor animal behaviours [14-18].

Farmers may more efficiently monitor the health and well-being of their animals by using these devices, which offer real-time data on the location, movement, and activity of individual animals. Early disease or distress signals can be identified with the help of this information, enabling timely intervention and treatment. Consequently, the ability of these smart devices to provide continuous insight into animal status has driven the widespread adaptation of various types of smart biosensor.

Among these, RFID-based sensors are particularly prevalent due to their comprehensive and robust capabilities in offering unique cow identification [19]. However, despite the growing advantages, the extensive reliance on tag readers and their susceptibility to environmental factors can accelerate the chances of damage or loss [10]. Conversely, GPS-based sensors are more affordable for tracking herds; however, their primary application is outside in pasture-based systems [13]. In contrast, Ultra-Wideband (UWB) indoor positioning systems present an important advancement for spatial behaviour monitoring, offering real-time animal location tracking with centimetre-level accuracy within complex barn environments [20, 21]. These cutting-edge monitoring systems can consistently track dairy cows without disrupting circadian rhythms [21], therefore overcoming challenges faced by other monitoring approaches. While UWB systems demonstrate strong prediction potential for analysing spatial patterns, behavioural changes, and welfare indicators when calibrated effectively [22], they are also susceptible to interference and damage from the barn environment, necessitating careful system deployment, data interpretation and maintenance [21, 23]. Similarly, IMUs, though not typically applied for object tracking, are frequently integrated with other localization devices to optimize overall performance and accuracy [24, 25]. While contact-based smart devices offer valuable insights, achieving comprehensive and precise individual monitoring of behavioural alterations/patterns such as eating and drinking bouts to aid early disease detection often presents challenges related to scalability, data continuity, integrity and environmental robustness. To overcomes these challenges and limitations, camera-based surveillance systems have emerged as a promising and powerful alternative tool for creating robust dairy cow monitoring systems [26]. These systems utilize surveillance cameras, commonly used to monitor farm conditions, to provide video data that can be analysed using computer vision (CV) algorithms to identify and track dairy cows [27]. The use of visual characteristics for dairy cow identification has grown in popularity because of CV technologies [28-31], while automatic cattle identification has made extensive use of machine learning (ML) and deep learning (DL) techniques [29, 30, 32-34].

## 2. Related works

An effective multi-object tracking system (MOT) for livestock would be a key platform for corporate dairy farm management, facilitating seamless monitoring of animal movements, health, and social interaction, ultimately contributing to enhanced efficiency and cost-effectiveness [35]. Monitoring dairy cows has grown difficult owing to alterations in barn conditions and the switch to a free-barn/free-stall rearing system. This amplifies the intricacy of behavioural analysis, since alterations in cow posture and shooting angles result in substantial variations in image characteristics. Multi-camera image analysis is essential to comprehensively capture the barn without any blind areas. Tracking dairy cows in a barn has been restricted to environments with single-camera [36, 37] set ups with limited flexibility. Multi-Target Multi-Camera Tracking (MTMCT) throughout a barn proves difficult considering to the subtle fluctuations in cows' appearance and posture, which prevent precise feature recognition.

A significant amount of progress has been made recently in tracking multiple animals at once, like cows, pigs, and sheep. Zheng et al. introduced a YOLO (You Only Look Once) Byte Track approach for cow tracking, leveraging the YOLOv7 framework and the Byte Track algorithm. They further reported a Higher Order Tracking Accuracy (HOTA) score of 67.60% by integrating the Self Attention and Convolution (ACmix) and Spatial Pyramid Pooling Cross Stage Partial Connections (SPPCSPC-L) models into the detector and fine-tunning Kalmin filter parameters [38]. Similarly, other researchers have improved the YOLOv5 model and DeepSort algorithms, amplifying them with improved ACmix and SPPCSPC-L modules to illustrate superior performance [27]. These outcomes have implicitly confronted challenges for instance, variations in cow sample size, erratic movements, and occlusions. The later authors further utilize a nonlinear motion model and a bench-matching mechanism, while attaining a HOTA score of 72.58%. Tu et al. developed and modified the TransTrack model for pigs, while attaining a HOTA score of 69.8% by magnifying the data association approach to minimize unnecessary tracking errors [39]. Guo et al. further improved three algorithms for tracking multiple objects: Joint detection and embedding (JDE), FairMOT, and YOLOv5s with DeepSort, and introduce a weight association method that smartly leverages the appearance embedded data[40].

Furthermore, Lu et al. proposed an oriented object detection approach to track multiple pigs from an overhead perspective [41]. This methodology minimizes background noise while introducing rotating bounding boxes and employing an enhanced Byte method to mitigate animal ID switches resulting from object movements and distortion during tracking. More recently, Zheng and Qin improved cow behaviour detection and monitoring by utilizing channel pruning to streamline the detector and Cascade Buffered Intersection over Union (C-BIoU) to optimize tracking performance, resulting in a HOTA score of 76.40% [42]. Likewise, for pig behaviour identification and tracking, Tu et al. used YOLOv5 and Byte Track, acquiring a HOTA score of 76.50% [43]. Nevertheless, in the context of intensive dairy farming, a recent lactation-based multi-camera model has reported tracking accuracies of approximately 90% for MOTA and 80% for IDF1 [26].

Building upon recent advances in computer vision (CV) and addressing the limitations observed in previous livestock tracking systems, this study proposes a unified, high-precision, multi-camera framework for real-time monitoring of indoor-housed dairy cows. Unlike earlier approaches that rely on single-camera setups, visual re-identification, or appearance-based embeddings, our method emphasizes geometric consistency and spatial continuity by integrating multiple overhead close circuit television (CCTV) feeds into a single panoramic view using homographic projection. This alignment eliminates the need for explicit cross-camera identity matching and enables seamless tracking across camera boundaries. To ensure both detection accuracy and instance-level segmentation fidelity, we combine an anchor-free YOLOv11-m detector with SAMURAI, a memory-enhanced, zero-shot segmentation model adapted from Segment Anything Model (SAM) 2.1. The key objectives of this research are:

- To design a panoramic multi-camera system that geometrically aligns overlapping CCTV views for complete barn coverage without redundant detections or identity mismatches.
- To fine-tune a high-performance object detector (YOLO11-m) for overhead cow detection and pair it with SAMURAI for generating temporally consistent, pixel-accurate masks.
- To evaluate and compare the tracking performance of the proposed system with existing baselines, specifically Deep SORT, under real-world conditions involving occlusion, overlapping animals, and variable postures.

3. Materials

*3.1. Data collection*

This research was conducted on indoor-housed Holstein Friesian dairy cows at the John Oldacre Centre for Dairy Welfare and Sustainability Research, a 200-head commercial dairy farm at Bristol Veterinary School, University of Bristol, United Kingdom. Six CCTV network cameras (IDs C16 to C21) were set up approximately 5 meters above the transition pen, positioned for top-down projection. Each camera was set to record at a resolution of 2592 x 1944 pixels. We recorded two distinct sets of footage, referred to as Video1 and Video2, 19 minutes and 59 minutes in duration respectively. Data were obtained six weeks apart (October 2024 and December 2024). As the transition pen is where cows are housed for approximately two weeks prior to calving, and is a self-contained pen with feeding, drinking and straw-based laying areas, The 14 cows in Video1 are different individuals than the 10 cows in Video2 (Section video-preprocessing).

## 3.2. Camera layout

To monitor an entire pen with six cameras, we first calibrated each camera to a common ground-plane coordinate system. A planar homography was computed for each camera using known correspondences on the pen floor, allowing us to warp and project each view onto a shared top-down plane. We cropped each transformed feed to its area of coverage and stitched them together into a near-seamless panoramic view of the pen with minimal overlap between camera regions. This preprocessing yielded a unified overhead view of the area of interest, simplifying multi-camera tracking by treating the merged feeds as a single expansive view. Overlapping fields of view were carefully managed (using slight cropping in overlap regions) to avoid duplicate detections of the same cow. By aligning all cameras in space, cows moving across camera boundaries remained in one continuous coordinate frame, eliminating the need for explicit re-identification across views.

## 4. Methodology

## 4.1. Video pre-processing

The preprocessing includes the following sequential operations, all implemented in Python (version: 3.10) using OpenCV and ffmpeg with unified device architecture (CUDA) acceleration on a Nvidia RTX 3090 GPU.

**Table 1 Video Pre-Processing Steps**

| Step | Operations | Description |
|---|---|---|
| 01 | Barrel distortion correction | Instead of relying exclusively on checkerboard-based calibration, we estimated barrel distortion coefficients directly from structural features present in the pen environment, such as the walls, wooden frames, and |
| | | partitions separating resting and feeding areas. These features, known to be straight in the real world, exhibited visible curvature due to lens distortion. We detected these lines using edge detection and then optimized distortion parameters by minimizing the deviation of the detected curves from their ideal straight forms. |
| 02 | Spatial cropping | All camera feeds were cropped manually in a way to minimize the overlapping area between frames. |
| 03 | Homographic wrapping | A planar projective transform (The matrix $H_k$ belongs to the set of 3×3 real matrices) was solved for each camera from a temporary $3 \times 3$ floor grid, yielding a root mean square (RMS) reprojection error of 1.5. Bilinear interpolation produced fully metric and bird's-eye views. |
| 04 | Temporal down sampling | Only six frames per-second were extracted from the original videos and merged at 30 fps, effectively shortening the total duration of both Videos from 19 and 59 minutes to 31 seconds for Video1 and 1 minutes and 59 seconds for Video2) and minimizing processing overhead. |
| 05 | Mosaic construction | The six rectified, down-sampled views were merged into a single 2224× 1084 canvas. Overlaps were minimized by (i) computing pairwise intersections, (ii) removing a 10 px feathered strip from the view contributing fewer valid pixels, and (iii) α-blending any residual overlap to avoid seams. |

*4.2. Proposed method*

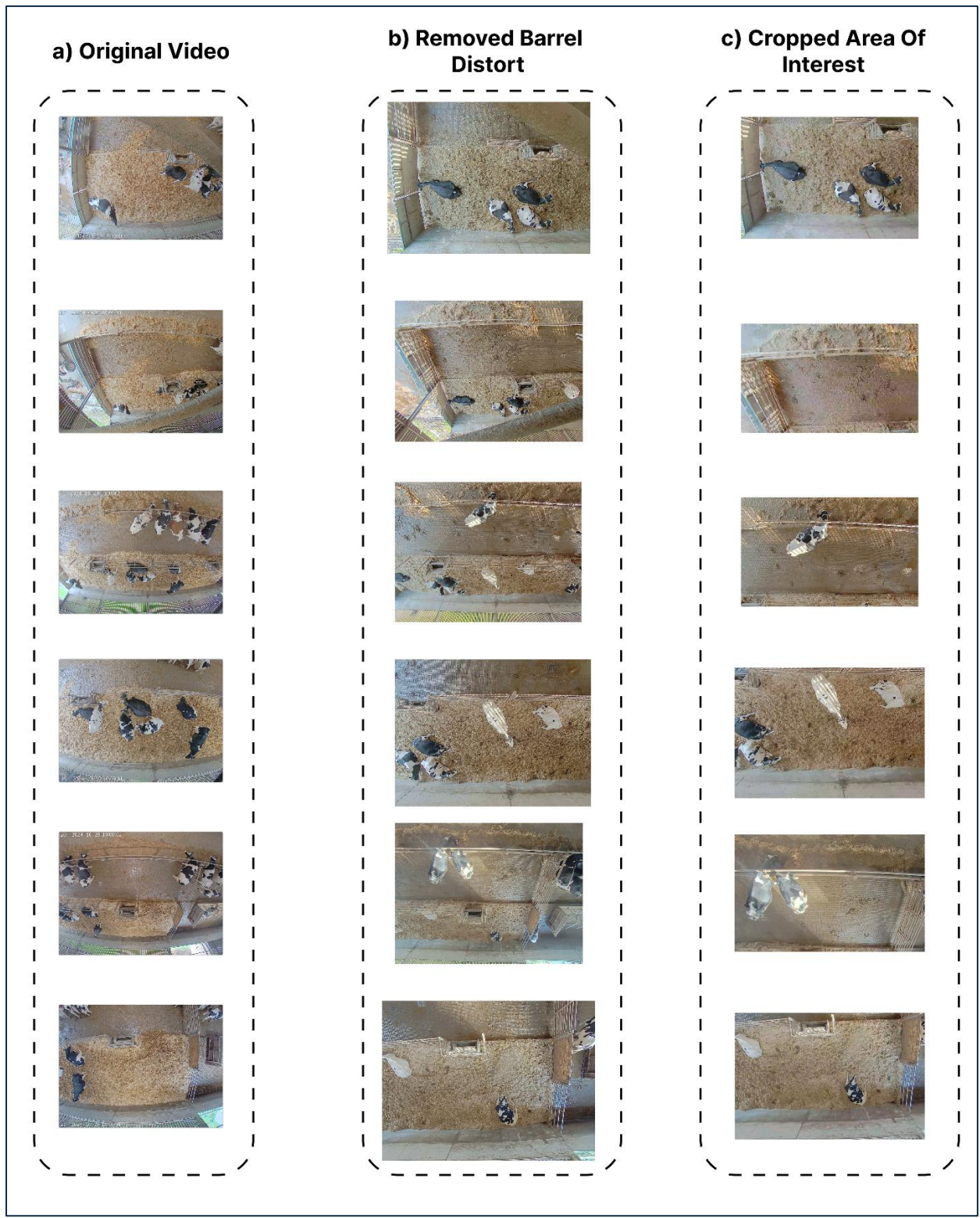

**Fig. 1. Our proposed method is based on a framework encompassing detection and tracking using SAMURAI.** Data feed from 6 cameras (a) were pre-processed by removing lens distortion

(b) and manually cropping (c) before stitching all camera feeds into one single video (See Video Pre-Processing Section).

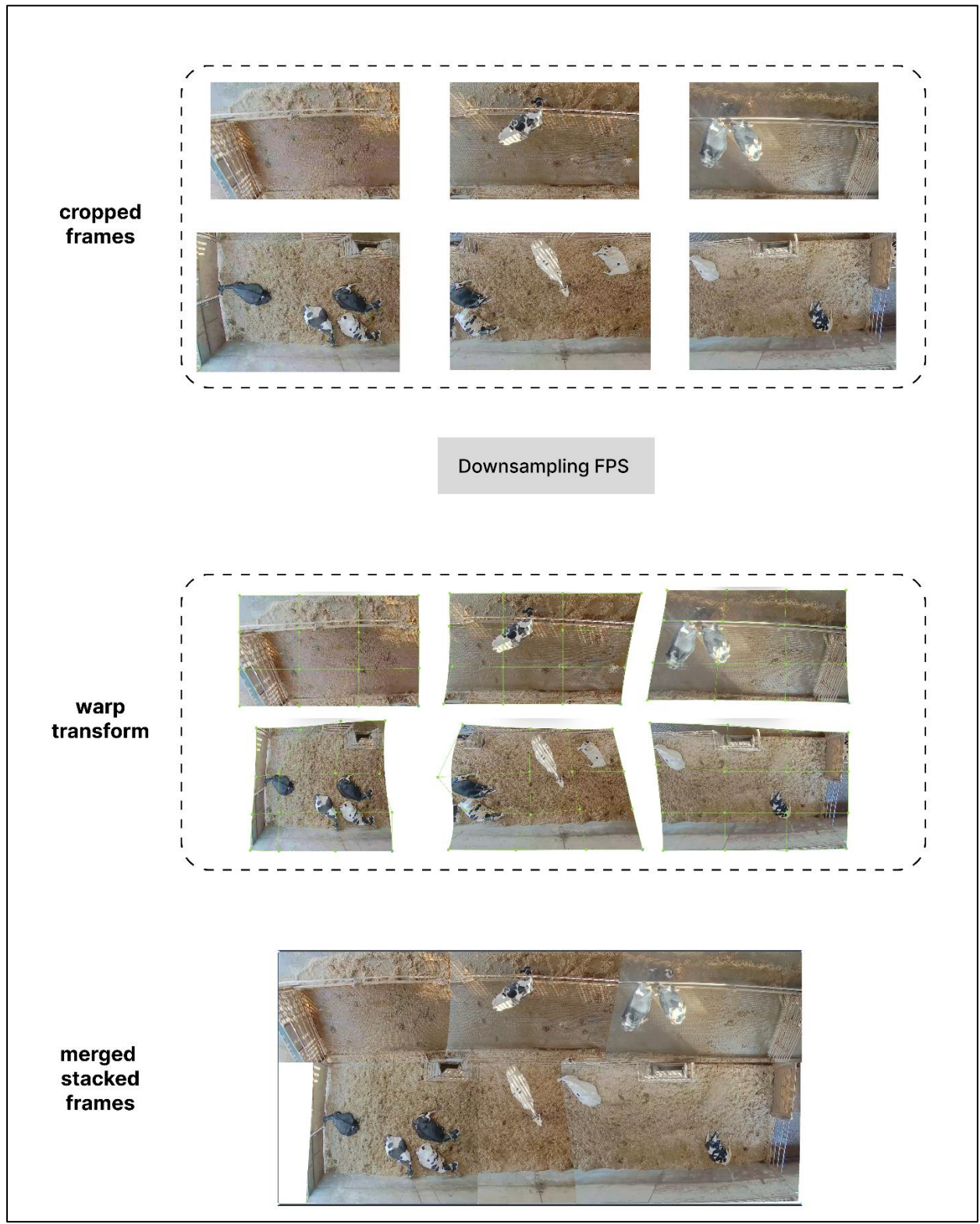

**Fig. 2.** The diagram depicts the spatial preprocessing and integration pipeline utilized for merging six camera feeds into one panoramic footage. Individual frames are first retrieved and down sampled, subsequently undergoing warp transformations for perspective and alignment correction, and ultimately combined into a cohesive composite picture of the monitored area.

### *4.3. Cow detection*

A custom dataset containing 2129 images was collected across all available cameras over a seven-month period, from June 2024 to January 2025, and 200 images were collected from the merged panoramic top view as Test dataset with a total sum of 2329 images (see supplementary data). The dataset encompassed a diverse range of lighting conditions, cow postures, and spatial configurations, ensuring broad representativeness of real-world scenarios. The frames were selected at random to avoid temporal or positional bias. A Ultralytics YOLO11 medium object detection model was fine-tuned with training and validation sets; the validation set supported hyperparameter tuning and early stopping. The test set remained separate for final performance evaluation only.

Each annotated image contained one or more cow instances, with bounding boxes manually drawn to label the location and extent of each animal present in the frame. The images were treated as independent observations, as the goal of the task was object detection (i.e., identifying the presence and location of cows in each frame), rather than individual recognition or identity tracking.

The dataset was split into three mutually exclusive subsets:

Training set: 1,895 images (approximately 81%),

Validation set: 234 images (approximately 10%),

Test set: 200 images (approximately 9%).

**Table 2** Cow detection dataset (Training and Validation) containing 2129 images was collected using single camera feed from different camera angles and lightening conditions. While 200 images were collected from the merged panoramic top view as Test dataset. As shown in table below.

| Training/Validation | |
|---|---|
| | 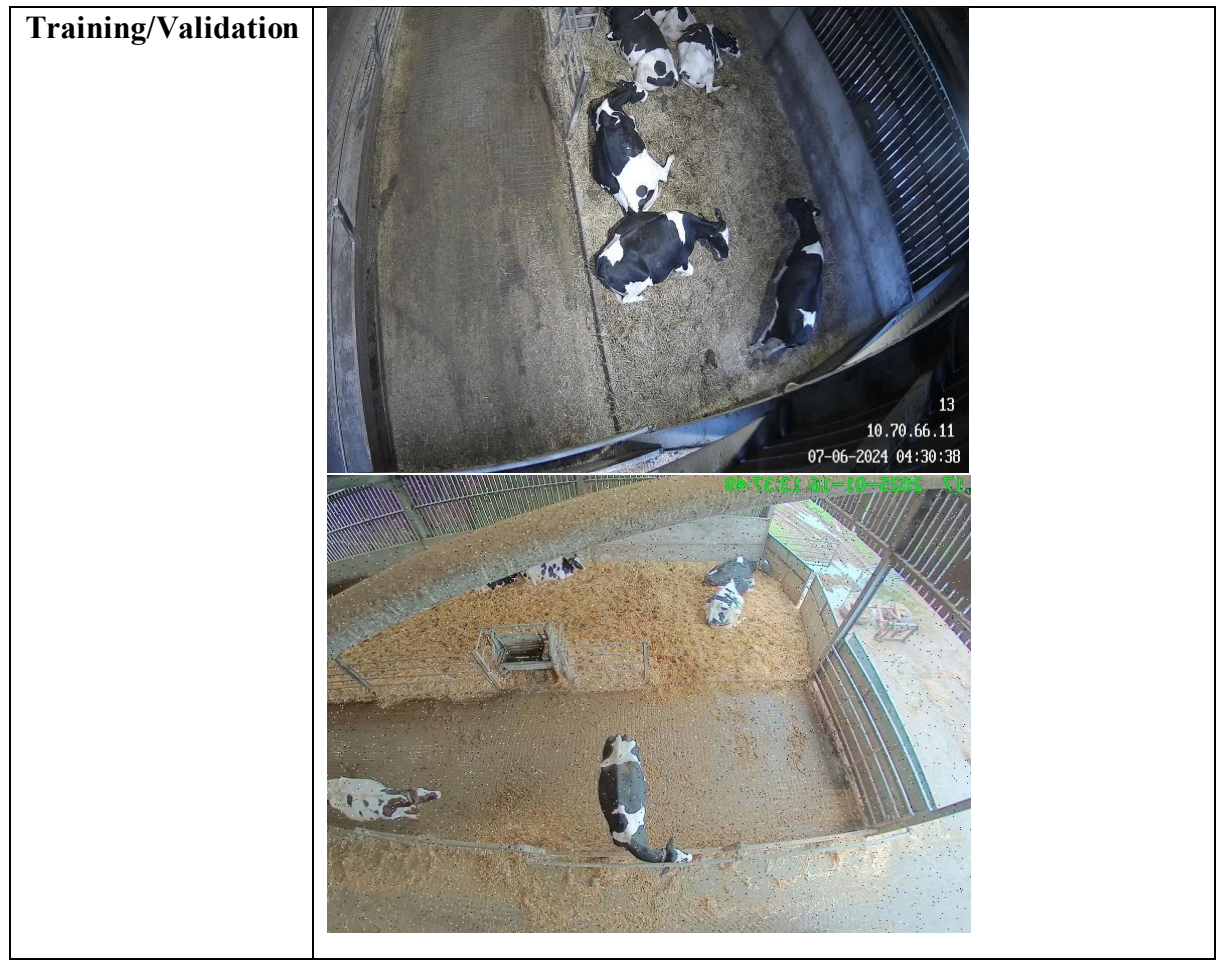 |

| Testing | 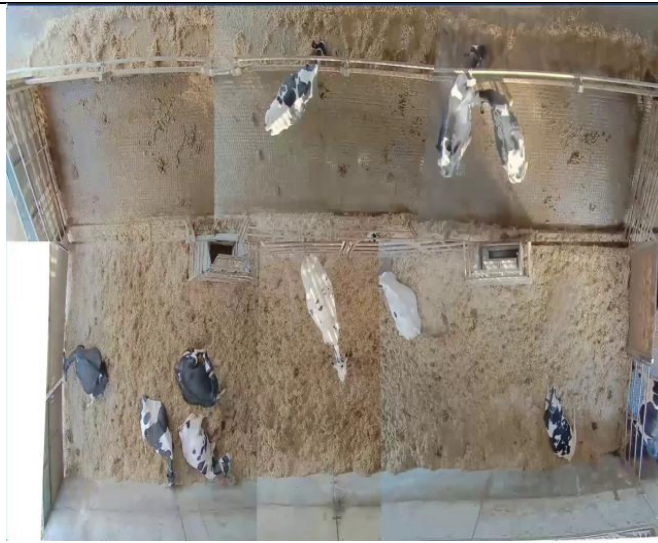 |
|---|---|

Training ran for 50 epochs with a batch size of 16 on 640 × 640 inputs. At inference time each 640 × 640 frame from the composite top-down view is processed in a single forward pass; confidence filtering (0.25) followed by NMS (IoU = 0.7) yields one tight bounding box per cow, maintaining robust performance under variable lighting and arbitrary animal orientation (Fig 3).

### 4.4. Instance segmentation with SAMURAI

To obtain precise per-cow masks and maintain temporal consistency across video frames, we used SAMURAI [44], an enhanced version of the Segment Anything Model 2.1 (SAM 2.1) [45] adapted for zero-shot visual object tracking via motion-aware memory. Each cow detected by our YOLO11-m model (via bounding box prompts) is handled by SAMURAI, which incorporates a Linear Kalman filter–based motion model and a hybrid memory-selection mechanism. This allows it to predict mask placement and choose high-quality historical frames when generating instance masks [44].

This pipeline refines the coarse object detections into accurate silhouettes while preserving object identity over time - even when cows are overlapping, occluding, or moving rapidly. SAMURAI operates in real time without any task-specific retraining, delivering true zero-shot performance.

We then derived updated bounding boxes from each segmentation mask by computing its minimal enclosing rectangle, providing both spatial precision and temporal coherence. These segmentation-enhanced boxes significantly reduced ambiguity in dense scenes, improving tracking reliability and reducing identity switches in crowded environments.

*4.5. Multi Object tracking*

At the outset of tracking (frame 0), each cow detected by YOLO11-m, a real-time object detector, is assigned a unique track ID. Each track's initial state is defined by the YOLO11-m bounding box and the corresponding segmentation mask produced by SAMURAI. During inference, our system uses a tracking-by-detection approach: for first frame, YOLO11-m detects cows and prompts SAMURAI to generate masks. And for the rest of the frames SAMURAI keeps segments for each detection across the video (See Graphical Abstract step 'd' and 'e'). We treated each mask $\mathbf{M_i}$ as an observation and tracked cows across frames with a two-stage algorithm:

1. **Motion prediction.** The centroid and area of every mask were propagated with a constant-velocity Kalman filter (state = [x, y, $v_x$, $v\_y$]).
2. **Data association.** At every frame $t$ we built a cost matrix **C** where $C_{ij} = 1 - \text{IoU}(*M_i, \mathbf{M_j})$. Linear assignment was solved via the Hungarian algorithm with a cut-off cost of 0.6. Unmatched tracks were kept alive for a maximum gap of 15 frames; newborn tracks required three consecutive hits before being confirmed.

Under the assumption of a closed population (no cows enter or leave), we maintained a fixed set of track IDs. As a result, identity ambiguity is eliminated: each mask matches at most one existing track, and no new track initiation is necessary. SAMURAI's precise instance segmentation, even under occlusions or cow crowding, further reinforces correct associations, reducing identity switches.

Finally, we updated track states each frame using the latest mask-derived bounding box. No appearance features or re-identification modules require spatial continuity, accurate segmentation, and motion-aware memory suffice for robust identity maintenance and stable trajectories.

### 4.5. Evaluation metrics

In this study we used four evaluation metrics; Precision (P), Recall (R), Mean Average Precision (mAP) and F1 score to evaluate the effectiveness of our detector model. The mathematical equations of these evaluations' metrics are described below.

$$Precision = TP/TP + FP \tag{1}$$

$$Recall = TP/TP + FN \tag{2}$$

Where;

TP = True positive.

FP = False positive.

FN = False negative.

$$Mean\ average\ precision(mAp) = \frac{1}{N}\sum_{i=1}^{N} AP_i \tag{3}$$

Where;

N = total number of classes.

$AP_i$ = Average precision for class, $i$.

$\sum_{i=1}^{N}$ = indicate the average precision from class 1 to class N.

To evaluate the performance of tracker, we used multi object tracking accuracy (MOTA), multi object tracking precision (MOTP), and IDF1 score (IDF1).

$$MOTA = 1 - FN + FP + \frac{IDSW}{GT} \qquad (4)$$

Where;

IDSW (Identity switches) = The number of times tracked object changes its identity (ID) in the output as compared to ground truth.

GT (Ground truth) = The total number of actual objects instances annotated in the dataset.

$$IDF1 = 2.IDTP/2.IDTP + IDFP + IDFN \qquad (5)$$

Where;

IDTP (Identity true positive) = The total number of correctly identified objects where detected identity matches with the ground truth (GT) identity of correct time span.

IDFP (Identity false positive) = The total number of detected objects that do not correspond to GT identity (false identities).

IDFN (Identity false negative) = The total number of GT identities/objects missed the tracker/model.

$$MOTP = \frac{\sum_{t,i} d_{t,i}}{\sum_t c_t} \qquad (6)$$

Where;

t = frame index in the sequence.

$c_t$ = the number of correct associations/matches at time *t*.

$d_{t,i}$ = the distance or intersection of union between the *i*-th matched hypothesis and its corresponding ground truth.

$\sum_{t,i}$ = Summation for all frames t, and for all frames, over all matched pairs, *i*.

$\sum_{t}$ = Summation for all frames t in the entire sequences.

## 5. Results and Analysis

### 5.1. Cow detection

The performance assessment of the YOLO11-m model for cows' detection during 50 training epochs is illustrated in Figure 3, displaying training and validation loss metrics with precision and recall-based accuracy assessments.

The training losses for bounding box (box_loss), classification (cls_loss), and distribution focal loss (dfl_loss) continuously decreased, indicating satisfactory convergence. Box_loss diminished consistently from an initial value of around 1.2 to 0.6, cls_loss decreased from about 1.0 to 0.4, and dfl_loss had a minor decline from 1.0 to 0.9. In contrast, validation losses exhibited significant variability, with notable peaks recorded in the earliest epochs (box\_loss \~4.0 and cls\_loss \~8.0). Nevertheless, these metrics stabilized markedly after roughly 10 epochs, converging to values approaching 1.0, underscoring initial learning difficulties that were later surmounted.

The evaluation criteria, encompassing precision and memory, indicated a swift and steady enhancement. Precision neared perfection (\~1.0), indicating exceptional accuracy in identifying actual cows with low false positives. Recall showed consistent enhancement, stabilizing at 0.9, signifying proficient identification of most actual cow cases in the sample.

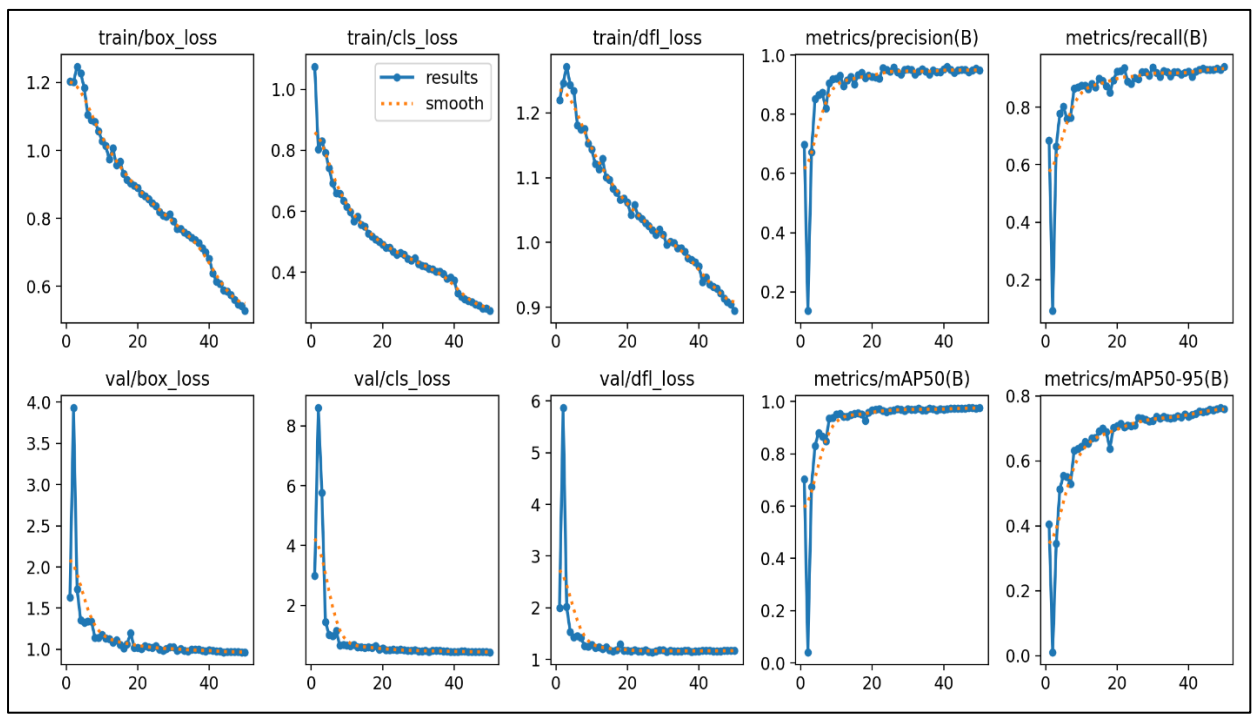

**Fig. 3. Evaluation metrics for YOLO11-m cow detection.** Training converged smoothly: all three loss terms fell monotonically, and the validation curves tracked the training curves closely, indicating negligible over-fitting. After 50 epochs the network reached a mean Average Precision (mAP) of **0.97 at IoU 0.5 (mAP50)** and **0.74 averaged over IoU 0.5–0.95 (mAP50–95)**. The optimum operating point, obtained from the F1–confidence sweep (Fig. 3), occurs at a confidence threshold of 0.454 and yields an **F1 = 0.94**, **precision = 0.96**, and **recall = 0.90**.

Mean Average Precision (mAP), assessed at an Intersection over Union (IoU) threshold of 50% (mAP50), attained near-optimal values (~1.0), so affirming outstanding detection efficacy at a widely utilized benchmark. Moreover, the severe mean Average Precision (mAP) assessed across Intersection over Union (IoU) thresholds from 50% to 95% (mAP50-95) shown strong localization skills, with values of 0.7, signifying precise and dependable detection performance even under more demanding assessment standards.

The results demonstrate that YOLO11-m effectively detects cows from image data, highlighting its potential utility in precision livestock management and monitoring applications. Subsequent research should assess model robustness across heterogeneous cattle populations and different climatic circumstances to improve its practical applicability.

*5.2. Instance segmentation*

Using the YOLO detections as point-prompts, SAMURAI (SAM2) returned high-quality masks. On the 400-image validation set the mean intersection-over-union was **0.926 ± 0.03**, confirming that the prompt strategy (centre-point positive, mid-edge negatives) reliably prevent mask leakage into the pen background while preserving fine anatomical detail needed for posture analysis.

*5.3. Multi-object tracking*

The performance evaluation clearly illustrates the superiority of the Proposed Method over Deep SORT Realtime in the context of multi-camera, multi-cow tracking. Across both video sequences, the Proposed method consistently achieved outstanding metrics: in Video1(SAM), it reached a Multi object tracking accuracy (MOTA) of 98.7% and in Video2(SAM), 99.3%, reflecting near-perfect tracking precision (Table 3). It maintained excellent identity preservation, with IDF1 scores exceeding 99%, and achieved highly accurate localization **Multi object tracking precision (MOTP)** as low as 0.001. The method demonstrated robustness by producing minimal false positives and false negatives (83–77), very few identity switches (18 in Video1(SAM) and only 52 in Video2(SAM)), and high detection accuracy (0.900 and 0.995 respectively). (See Supplementary Video1(SAM) and Supplementary Video2(SAM)).

**Table 3 Proposed Method using SAMURAI (SAM2.1):** Here; MOTA indicates the multi-object tracking accuracy and was calculated with equation: $MOTA = 1 - FN + FP + \frac{IDSW}{GT}$; MOTP represents as multi-object tracking precision and calculated as follows: $MOTP = \frac{\sum_{t,i} d_{t,i}}{\sum_t c_t}$; IDF1 demonstrate the identification F1 score and was calculated as: $DF1 = 2.IDTP/2.IDTP + IDFP + IDFN$; FP = False positive, FN = False negative, IDs = represent the identification numbers, and DetA= detection accuracy and was calculated with equation: $DetA = True\ positive/True\ positive + False\ negative + False\ positive$. Seq-A and Seq-B represent Supplementary Video1 and Video2 respectively.

| Sequence | Frames | MOTA ↑ | MOTP ↓ | IDF1 ↑ | FP | FN | IDs | DetA ↑ |
|---|---|---|---|---|---|---|---|---|
| Video1(SAM) | 944 | 98.7% | 0.001 | 99.3% | 77 | 77 | 18 | 0.900 |
| Video2(SAM) | 3333 | 99.3% | 0.003 | 99.7% | 83 | 83 | 52 | 0.995 |

In stark contrast, Deep SORT Realtime showed notably poor performance, especially in Video2(DeepSort), where it recorded a negative MOTA of -14.1% and an extremely low IDF1 of 12.6%, alongside an overwhelming number of false detections and identity mismatches. Even in Video3, Deep SORT's performance remained suboptimal, with a MOTA of only 48.1% and an IDF1 of 22.6% (Table 4). Notably, on the longer 3,333-frame sequence (Video2(SAM)), the Proposed Method missed or falsely introduced fewer than 0.25% of objects, and identity switches occurred in only 0.16% of cases, while maintaining over 99% detection accuracy. Its consistent performance even in the more complex Video1(SAM) highlights its robustness to variations in camera angle and lighting. These results strongly affirm the reliability and effectiveness of the Proposed Method for high-precision livestock monitoring in complex environments. (See Supplementary Video1(DeepSort) and Supplementary Video2(DeepSort)).

**Table 4 Deep-Sort Realtime:** Here; MOTA indicates the multi-object tracking accuracy and was calculated with equation: $MOTA = 1 - FN + FP + \frac{IDSW}{GT}$; MOTP represents as multi-object tracking precision and calculated as follows: $MOTP = \frac{\sum_{t,i} d_{t,i}}{\sum_t c_t}$; IDF1 demonstrate the identification F1 score and was calculated as: $DF1 = 2.IDTP/2.IDTP + IDFP + IDFN$; FP = False positive, FN = False negative, IDs = represent the identification numbers, and DetA= detection accuracy and was calculated with equation: $DetA = True\ positive/True\ positive + False\ negative + False\ positive$. Seq-A and Seq-B represent Supplementary Video3 and Video4 respectively.

| Sequence | Frames | MOTA ↑ | MOTP ↓ | IDF1 ↑ | FP | FN | IDs | DetA ↑ |
|---|---|---|---|---|---|---|---|---|
| Video1(DeepSort) | 944 | 48.1% | 0.117 | 22.6% | 838 | 838 | 4997 | 80.71% |
| Video2(DeepSort) | 3333 | -14.1% | 0.194 | 12.6% | 16475 | 16475 | 4730 | 26.38% |

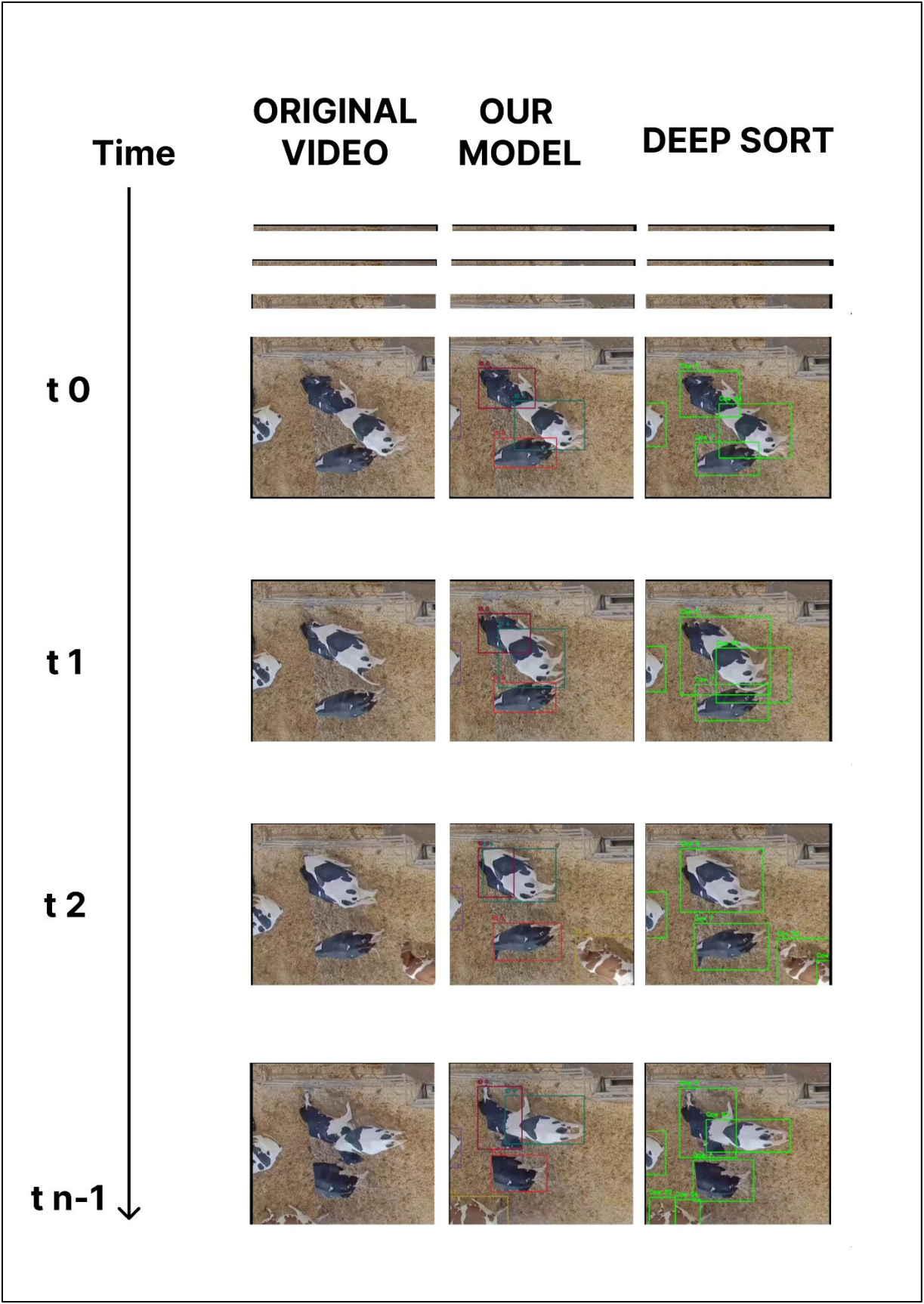

**Fig. 4. Comparative analysis of object tracking performance over time in complex multi-cow scenarios using SAMURI (proposed model) versus Deep SORT:** The figure illustrates tracking consistency across four time points ($t_0$ to $t_{n-1}$). The left column presents the original unannotated video frames. The middle column shows tracking results using the proposed SAMURI model, while the right column shows outputs from the baseline Deep SORT tracker. SAMURI maintains tighter, more consistent bounding boxes with less ID switching and improved spatial precision across frames, particularly under occlusion and overlapping instances. In contrast, Deep SORT exhibits multiple ID reassignments and bounding box jitter, especially as cow positions shift.

## 6. Conclusion

By integrating all camera feeds into a single homography-aligned view, our proposed tracking system inherently ensures seamless continuity for dairy cows transitioning between various camera coverages. For instance, when a cow walks from the area of Camera A into Camera B's coverage, its track persists seamlessly. The homography alignment means the cow's image simply moves across the merged frame, and SAMURAI continues tracking it as the same object (with YOLO11-m detection for first frame) in the new camera's region. This avoids the common problem in multi-camera systems of mistakenly treating the same cow as a different individual when it appears in another camera. Our approach eliminates the need for separate cross-camera re-identification algorithm or appearance matching; instead, the continuity is ensured by the unified coordinate space and consistent instance segmentation. It is worth noting that this strategy assumes each cow is visible in at most one camera at a time a condition ensured by our camera placement within minimal overlap. Consequently, the proposed methodology prevents duplicate detections of individual cow across multiple cameras and ensure that each tracking ID remain unique throughout the entire pen

Through reliance solely on spatial features and motion-aware memory, the proposed method avoids the complexity and brittleness of appearance-based trackers, offering significant improvements in tracking accuracy (MOTA), identity preservation (IDF1), and detection robustness. The system demonstrates strong generalizability, minimal identity switching, and superior performance in complex indoor farm environments, laying the groundwork for future behavioural modelling and health monitoring applications.

## 7. Limitations and future work

Despite the above idealized performance, our system has some practical limitations. Closed-world assumption: The tracker assumes a fixed set of cow identities initialized at the start. It cannot handle new cows introduced post-initialization – if an untracked cow enters the pen, the system currently has no mechanism to assign it a new ID. In such a case, the new cow's detections would remain unmatched to any existing track, leading to untracked objects (which would count as false negatives) or would require manual reinitialization of the tracking system. Similarly, if a cow leaves the pen or is occluded for a long duration, the system will not gracefully handle reintroducing it, since it relies on continuous presence and spatial continuity. Another limitation is the reliance on accurate homography calibration; significant calibration errors could cause misalignment between camera views, potentially breaking the seamless tracking when cows cross camera boundaries. Additionally, the current approach does not explicitly model occlusions or long-term lost-and-found scenarios – though SAM's masks help with short occlusions, a prolonged disappearance of a cow (e.g. lying in a blind spot) might cause the track to be permanently lost. Future work will concentrate on various keys aspects. This includes scaling the system to accommodate a large number of dairy cows and extending the data collection period long term to enable robust longitudinal behavioural monitoring of individual dairy cow activity. Such classification will facilitate a panoramic daily activity monitoring paradigm, which is key for early disease detection.

**Authorship contribution**

Conceptualization and methodology, Kumail Abbas, Zeeshan Afzal, Ali Alameer, Chaidate Inchaisri; writing—original draft preparation, Kumail Abbas, Aqeel Raza; supervision, Taha Mansouri, Ali Alameer, Andrew Dowsey, Chaidate Inchaisri; project administration, Ali Alameer, Andrew Dowsey; funding acquisition, Chaidate Inchaisri.

**Funding**

Kumail Abbas and Chaidate Inchaisri were funded by The Second Century Fund (C2F), Chulalongkorn University Bangkok, Thailand, and the 90th Anniversary of Chulalongkorn University Ratchadaphisek Somphot Endowment Fund. The data collection and Andrew Dowsey's contribution is funded by BBSRC grant BB/X017559/1.

**Acknowledgment**

The authors extend their gratitude to Richard Bruce and Axel Montout at the John Oldacre Centre for Dairy Welfare and Sustainability Research, Bristol Veterinary School, for providing the data necessary for conducting the study.

**Conflicts of Interest**

The authors declare no conflicts of interest.

**Data availability**

The authors do not have permission to share data publicly.

**Supplementary data**

[Multi Camera Cows Tracking Results](#)